\documentclass{article}
\usepackage{microtype}
\usepackage{graphicx}
\usepackage{booktabs}
\usepackage{multirow}
\usepackage{enumitem}
\usepackage{placeins}
\usepackage{hyperref}

\usepackage[preprint]{icml2027}
\usepackage{amsmath}
\usepackage{amssymb}
\usepackage{mathtools}
\usepackage{amsthm}
\theoremstyle{plain}
\newtheorem{proposition}{Proposition}
\theoremstyle{definition}
\newtheorem{definition}{Definition}

\icmltitlerunning{What Is a Simulated Respondent Worth?}

\begin{document}

\twocolumn[
  \icmltitle{What Is a Simulated Respondent Worth?\\Stretching Survey Budgets with Language Models}

  \begin{icmlauthorlist}
    \icmlauthor{Geonwoo Bang}{eth}
    \icmlauthor{Dongho Kim}{skku}
    \icmlauthor{Juyoung Seo}{skku}
    \icmlauthor{Moohong Min}{skku}
  \end{icmlauthorlist}

  \icmlaffiliation{eth}{ETH Zürich, Zürich, Switzerland}
  \icmlaffiliation{skku}{Sungkyunkwan University, South Korea}

  \icmlcorrespondingauthor{Moohong Min}{iceo@skku.edu}

  \icmlkeywords{language models, human simulation, survey research, effective sample size, forecast combination}

  \vskip 0.3in
]

\printAffiliationsAndNotice{}

\begin{abstract}
High-quality surveys are expensive, and language models are increasingly used as simulators to stretch fieldwork budgets. To determine where they save interviews, we measure any simulator in units of real respondents: a simulator is worth $r$ respondents if its predictions match the average accuracy of $r$ random people. Under optimal weighting, pooling $n$ independent real respondents with a simulator worth $r$ yields an estimate worth $n+r$. Across 961 questions from seven survey programmes, sixteen simulators from six developers are worth 6 to 23 respondents, and 7 to 49 on 250 post-training poll questions. Combining simulators saves almost nothing because they share most errors: all sixteen reach 25 respondents, barely above the best single model (23). Real answers from the same survey help far more, as within-model error overlap lets known answers cancel errors on related items. For the strongest simulator, twelve other models add only 1.5 respondents, whereas prompting with real results from three other questions adds 7, and half the survey adds 19. For a demographic or political subgroup, the strongest simulator alone is worth 27 group respondents and the overall result 15; prompted with that overall result, it reaches 163. All findings replicate on two held-out later poll reports, yielding a playbook for stretching survey budgets alongside real respondents.
\end{abstract}

\section{Introduction}
\label{sec:intro}

High-quality surveys are costly, especially for long questionnaires or small subgroups. Language models offer a way to stretch survey budgets: given a population description and a question, a model estimates the share choosing each option \citep{argyle}. Such \emph{simulators} are used to pretest questionnaires, fill in under-sampled subgroups, and sometimes replace surveys altogether.

Using simulators to save money raises concrete budgeting questions: how many interviews does a simulator save, where should limited interviews go, and whether extra budget helps more on additional models or real answers? Existing evaluations report abstract distribution distances \citep{opinionqa,simbench}, yet a distance of 0.02 does not tell a practitioner how many respondents it replaces or how to redesign a survey.

We place simulators and fieldwork on one budgeting currency: real respondents. A simulator is \emph{worth $r$ respondents} if its predictions match the average accuracy of $r$ random people from the population (Section~\ref{sec:method}). This unit is intuitive to anyone running a survey and additive under optimal weighting: combining $n$ independent real respondents with a simulator worth $r$ yields an estimate worth $n+r$ respondents. Placing other sources, such as a second simulator, an earlier poll, or other survey answers, on the same scale lets us allocate limited interviews where they help most.

We evaluate sixteen simulators from six developers, from 7-billion-parameter open models to frontier proprietary systems, across 961 questions from seven survey programmes and roughly 450 post-training poll questions. Three findings govern how to spend a survey budget.\footnote{Code: \href{https://github.com/g7199/simulated-respondent-worth}{github.com/g7199/simulated-respondent-worth}}

\begin{itemize}[leftmargin=*,itemsep=2pt,topsep=2pt]
\item \textbf{A simulator saves tens of respondents, not hundreds.} Individual simulators are worth 6 to 23 respondents on older surveys and 7 to 49 on new polls, with larger, higher-tier models scoring higher.
\item \textbf{Paying for more simulators adds little.} Simulators share most errors: adding a second model raises a stronger model's worth by a median of 2.7\%. With orthogonal errors, the sixteen models would sum to 207 respondents; combined, they reach only 25, barely above the best single model (23). This overlap imposes a measurable ceiling on non-negative ensembles.
\item \textbf{Real answers multiply a simulator's value.} On weekly polls (across thirteen direct-output simulators), twelve extra models add only 1.5 respondents to the strongest simulator, whereas prompting with real results from three other survey questions adds 7, and half the survey adds 19 (Figure~\ref{fig:real}a). For a demographic or political subgroup, the strongest simulator alone is worth 27 group respondents and the overall result 15; given that overall result in its prompt, the simulator jumps to 163.
\end{itemize}

Error overlap explains both patterns (Equation~\ref{eq:overlap}). Across models on the same question, shared errors fail to cancel when averaged; within one model across related questions, a real answer to one item cancels much of the error on another. All three findings replicate on two held-out later poll reports after fixing our protocol.

Together, these results yield a six-point playbook for stretching a survey budget (Section~\ref{sec:real}): invest in one strong model; cover every question thinly from the start; when omitting items, select representative questions for the prompt; solve subgroup estimation via accurate overall results; deprioritize questions with recent waves; and protect human samples for underrepresented groups and high-stakes items.

\section{Measuring a Simulator in Respondents}
\label{sec:method}

\paragraph{One question, one population.} Consider a survey question with $K$ answer options, where a true population share $\theta_k$ would choose option $k$, forming a distribution $\theta=(\theta_1,\dots,\theta_K)$. Drawing a simple random sample of $n$ people yields observed shares $\bar x$, whose expected squared Euclidean sampling error is
\begin{equation}
\mathbb{E}\,\|\bar x-\theta\|^2=\frac{V(\theta)}{n},\qquad V(\theta)=\sum_{k=1}^K\theta_k(1-\theta_k).
\label{eq:sample}
\end{equation}
Here $V(\theta)<1$ is the total variance of a single respondent's one-hot answer; for a 60/40 split ($V=0.48$), 100 respondents give roughly 4.9 percentage points RMSE per option, and quadrupling $n$ halves it.

Given the question and a population description, a \emph{simulator} returns predicted shares $c$ with error $c-\theta$. Because a single question cannot reliably judge a simulator, we average across a set of questions: let $G=\mathbb{E}\,\|c-\theta\|^2$ be the simulator's mean squared error and $V=\mathbb{E}\,V(\theta)$ the average single-respondent variance.

\begin{definition}[Worth in respondents]
A simulator is \emph{worth $r$ respondents} if its mean squared error equals that of $r$ random respondents, $G=V/r$; equivalently, $r=V/G$.
\end{definition}

The same definition applies to any \emph{source} of predicted shares, such as last month's answer to the same question, a neighbouring country's answer, or the overall population result when targeting a subgroup.

Worth summarizes average squared error only. Whereas random sampling error shrinks as more people are polled, a simulator's error on a given question is fixed and often correlated across related items. Thus $r$ characterizes a simulator over a class of questions rather than guaranteeing accuracy on any single item (Section~\ref{sec:ladder}), and must be estimated on similar questions (Section~\ref{sec:models}).

\paragraph{Combining a simulator with respondents.} The simplest way to combine $n$ real respondents with a simulator is a weighted average that treats the simulator's prediction $c$ as $w$ extra respondents:
\begin{equation}
\hat\theta=\frac{n\,\bar x+w\,c}{n+w}.
\label{eq:pool}
\end{equation}
When the $n$ real respondents are sampled independently of the simulator, setting $w=r$ is optimal and yields an estimate worth exactly $n+r$ respondents. Fortunately, the weight need not be exact: any positive $w$ up to $2r$ strictly beats ignoring the simulator, and even over-weighting at $w=3r$ increases mean squared error by at most 12.5\% relative to the respondents alone (Appendix~\ref{app:proof}).

\paragraph{Combining several sources.} With multiple sources, each source $j$ receives weight $w_j$, and their combined worth depends on whether they make the same errors. Write $e_j=c_j-\theta$ for source $j$'s error, and let $G$ be the uncentered error second-moment matrix with entries $G_{ij}=\mathbb{E}\langle e_i,e_j\rangle$ (so diagonal entries satisfy $G_{jj}=V/r_j$, while off-diagonal entries measure shared error). Let $\mathbf{1}$ denote the all-ones vector.

\begin{proposition}
\label{prop:main}
Suppose $G$ is positive definite and the real respondents are sampled independently of the sources. The pooled estimate $(n\bar x+\sum_j w_jc_j)/(n+\sum_j w_j)$ has mean squared error $(nV+w^\top G\,w)/(n+\mathbf{1}^\top w)^2$. Regardless of $n$, the optimal weights are $w^\star=V\,G^{-1}\mathbf{1}$, achieving a worth of exactly
\begin{equation}
n+r_S\quad\text{respondents},\qquad r_S=V\,\mathbf{1}^\top G^{-1}\mathbf{1}.
\label{eq:set}
\end{equation}
\end{proposition}

We call $r_S$ the worth of the set; Equation~\ref{eq:set} recasts classical optimal forecast combination \citep{batesgranger} in respondent units. In principle, $w^\star$ can place negative weights on weaker models to extrapolate; to avoid overfitting, our empirical ensembles restrict to non-negative weights $w\ge 0$, which yields nearly identical worth (Appendix~\ref{app:proof}). Two limiting cases show how error overlap governs whether extra models are worth paying for:

\emph{Orthogonal errors.} If errors are orthogonal ($G_{ij}=0$ for $i\ne j$), worth adds up ($r_S=\sum_j r_j$): ten simulators worth 20 each combine to 200.

\emph{Nested errors.} If a weaker source makes every error the stronger source makes plus unrelated noise ($e_{\text{weak}}=e_{\text{strong}}+\text{noise}$), then
\begin{equation}
G_{ij}=\min(G_{ii},G_{jj}),
\label{eq:nested}
\end{equation}
all optimal weight falls on the strongest source, and ten simulators worth 20 each remain worth 20. We call such errors \emph{nested}, a form of forecast encompassing \citep{chonghendry,hln}.

Between these extremes, error overlap caps the worth of any non-negative ensemble, reflecting classical limits on dependent information sources \citep{clemenwinkler}.

\begin{proposition}
\label{prop:ceiling}
Let $r_{\max}$ be the worth of the strongest source. If every pair $i,j$ satisfies $G_{ij}\ge\gamma\,\min_k G_{kk}$ for some $\gamma\in(0,1]$, then no combination with non-negative weights can exceed $r_{\max}/\gamma$ respondents, however many sources it includes.
\end{proposition}

We define the error correlation between two sources as $\rho_{ij}=G_{ij}/\sqrt{G_{ii}G_{jj}}$ (the uncentered correlation, natural here because option errors sum to zero on every item). When errors have equal size, $\gamma$ is simply the minimum pairwise error correlation and can be estimated from data. For any pair of sources, we also report the \emph{gain of a pair} (how much combining both beats the stronger source alone) and the \emph{overlap ratio} $G_{ij}/\min(G_{ii},G_{jj})$, which equals 0 for orthogonal errors and 1 for nested errors.

\paragraph{Using real answers.} Often real answers are already at hand for related quantities: other questions in the same survey, or the \emph{overall result} across all respondents when targeting a subgroup. We compare querying a simulator \emph{cold} (with no real answers) against placing those real answers in its prompt or applying a post-hoc correction. For a related target sharing the same response options, the simplest correction is an \emph{anchor}: given predictions $c$ for the target and $c'$ for a related quantity whose true answer $\theta'$ is known, the anchor is $\theta'+(c-c')$. Subtracting $c'$ from $c$ cancels any error shared across both predictions.

\paragraph{Why overlap cuts both ways.} Take two equal-sized errors with correlation $\rho$. If they come from two models on the same question, averaging them multiplies one model's worth by the first factor below. If they come from one model on two related questions, anchoring on one known real answer multiplies worth by the second factor:
\begin{equation}
\underbrace{\frac{2}{1+\rho}}_{\text{second model}}\qquad\text{and}\qquad\underbrace{\frac{1}{2(1-\rho)}}_{\text{real anchor}}.
\label{eq:overlap}
\end{equation}
High overlap hurts ensembling but helps anchoring: at $\rho=0.8$, a second model adds only 11\%, whereas a real anchor multiplies worth by 2.5.

\section{Data and Simulators}
\label{sec:setup}

\paragraph{Questions.} We evaluate simulators on three complementary sets of survey questions.

\emph{Older surveys (broad coverage).} These comprise 961 questions from seven large survey programmes: the General Social Survey (GSS), OpinionQA, the European Social Survey, Afrobarometer, ISSP, Latinobar\'ometro, and GlobalOpinionQA \citep{simbench,opinionqa,globalqa}. Each item targets a national population (so a question asked in two countries counts twice). Because these answers are public and predate the models, models may have seen some of them.

\emph{New polls (post-cutoff test).} To mitigate concerns about memorization of the survey results, we collected 250 questions from 25 poll releases by nine pollsters in nine countries, all conducted between 1 October and 6 November 2026, after every model was trained.

\emph{Weekly polls with groups (real-data and subgroup test).} From five weekly reports of one U.S.\ poll (The Economist/YouGov) in the same period, we extracted 246 distinct questions reporting full-population and twenty subgroup results (across demographics, 2024 vote, ideology, and party; Figure~\ref{fig:real}c), along with group sample sizes, yielding 4,920 group distributions in all. Forty-four questions also appear in the 250 new polls, giving 452 distinct post-cutoff questions overall.

\paragraph{Simulators.} We test sixteen language models from six developers: eleven proprietary GPT and Claude models and five open models (7B to 27B parameters) runnable on one or two consumer GPUs (Figure~\ref{fig:models}a). Every model receives the same request: a description of the respondents, the question and its options, and an instruction to estimate the percentage choosing each option. Models answer without tools or search, in batches of twenty questions unless noted otherwise. We use stated percentages \emph{as given}, except for the three smallest open models: because they struggle to output calibrated percentages directly, we prompt them as an individual respondent and extract next-token probabilities over options, rescaled with two out-of-sample constants (Appendix~\ref{app:details}).

\paragraph{Estimation and uncertainty.}
We compute $V$ and $G$ directly from published survey shares and model predictions, averaged across questions. Because every simulator is scored against the same finite-sample poll, the poll's sampling noise inflates both each simulator's apparent error and the error shared across simulators. For the weekly polls and their subgroups, where sample sizes are reported and subgroup samples can be small, we subtract this expected sampling variance from $G$ (Appendix~\ref{app:context}).

To avoid overfitting, we fit ensemble weights and post-hoc corrections on separate questions; whenever we instead fit them in-sample to give an optimistic upper bound, we say so explicitly. Brackets show 95\% bootstrap intervals, resampled over question texts (or over poll releases for the new polls). All three findings follow one uniform protocol, hold across protocol variants (Appendix~\ref{app:details}), and replicate once on two later poll reports held out until the protocol was frozen (Section~\ref{sec:real}).

\section{A Standalone Simulator Saves Tens of Respondents}
\label{sec:ladder}

\begin{figure*}[t]
\centering
\includegraphics[width=0.95\textwidth]{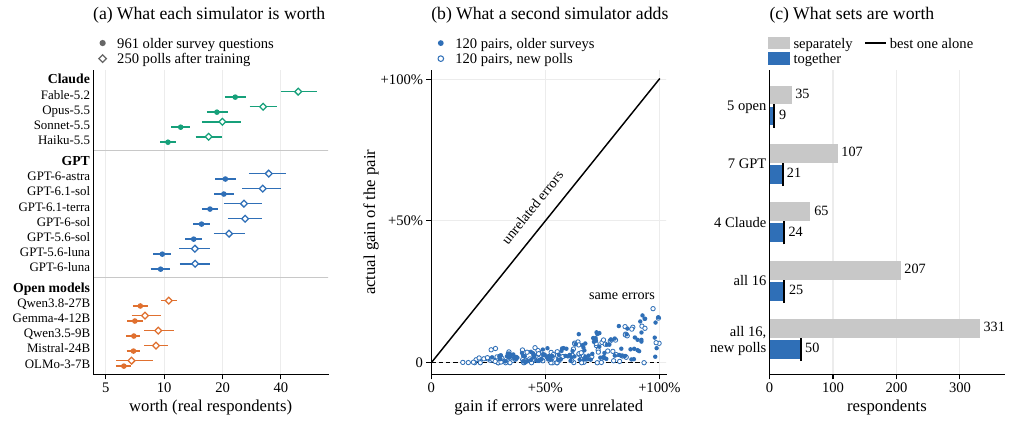}
\caption{(a) Worth of each simulator (in respondents, log scale), with 95\% intervals, on older surveys (dots) and new polls (diamonds). (b) Pairwise combinations across all 120 simulator pairs: expected gain over the stronger member if errors were orthogonal (horizontal) versus actual gain with in-sample non-negative weights, which overstates benefit (vertical). (c) Worth of simulator sets counted independently (grey), combined together with in-sample non-negative weights (blue), and for the strongest single model alone (black line).}
\label{fig:models}
\end{figure*}

Figure~\ref{fig:models}a shows how many interviews each simulator saves on its own. On older surveys, the five open models (7B to 27B) are worth 6 to 8 respondents, rising through mid-tier models (10 to 12) and upper-tier models (14 to 20) to GPT-6-astra and Claude-Fable-5.2 at 21 and 23. Larger and higher-tier models consistently score higher \citep{ziems}, yet even the best model matches only about two dozen random respondents. Three zero-cost baselines set the scale (Table~\ref{tab:ladder}): equal shares are worth 4 respondents, the average real answer to other questions with the same options is worth 8, and another population's real answer to the same question is worth 10, beating every open model.

Unusually high worth on widely cited older surveys suggests possible memorization: Claude-Fable-5.2 reaches 345 respondents on the 21 GSS questions, versus 13 to 79 across the other six programmes. The 250 post-cutoff polls mitigate memorization while preserving the same ranking (rank correlation 0.98), with worths of 7 to 49 respondents. Their higher worth tracks geography: new polls come mainly from the United States and Western Europe, where every simulator performs better (Table~\ref{tab:prog}).

Crucially for survey budgeting, standalone accuracy is highly uneven across items and populations. On three out of four questions, the best simulator beats 23 respondents, yet its worst tenth of questions causes half of its squared error. Worth also falls sharply for populations underrepresented online, such as Afrobarometer, Latinobar\'ometro, and GlobalOpinionQA (13 to 17 respondents even for the best model; Table~\ref{tab:prog}). A standalone simulator therefore cannot safely replace fieldwork on single high-stakes items or underrepresented groups.

Finally, more expensive prompting setups buy little extra accuracy (Appendix~\ref{app:details}). Simulating a panel of sixteen individual personas does not consistently beat direct share estimation, and rescaling or rewording changes worth only slightly. Querying one question per call instead of twenty helps the two tested Claude models by 9\% to 14\%, but not GPT models.

\section{More Simulators Buy Almost Nothing}
\label{sec:models}

If one simulator saves 23 respondents, can paying to run sixteen simulators together save much more? By Proposition~\ref{prop:main}, the answer depends on whether they make different errors.

\paragraph{A second simulator adds little.} Combining any two of our sixteen simulators (120 pairs) with optimal in-sample non-negative weights gives an upper bound on ensembling gain. Under orthogonal errors, a second model would boost the stronger member by 27\% to 100\%. Actual gains are far smaller (Figure~\ref{fig:models}b): the median gain is 2.7\%, nine pairs in ten gain 10\% or less, and the maximum is 17\% (median 1.7\%, maximum 19\% on new polls). Since averaging two repeat runs of the same model already adds 1\% to 3\%, a typical second model buys no more than querying the first model twice.

The reason is that errors are largely nested (Equation~\ref{eq:nested}): across all pairs, the pooled overlap ratio is 0.80 [0.77, 0.83]. Even cross-developer GPT--Claude pairs of comparable strength (within 17\% in worth) correlate 0.68 to 0.81 and gain only 5\% to 15\%.

\paragraph{Sixteen models barely beat one.} Counted independently, the sixteen simulators sum to 207 respondents; combined with optimal in-sample non-negative weights, they reach only 24.8 [22.0, 28.3], barely above the best single model at 23.2 (Figure~\ref{fig:models}c; 25.2 allowing negative weights). On new polls, 331 counted independently collapse to 50 combined versus 49 for the best model alone. More flexible combiners fitted out-of-sample on six programmes reach at most 24.4 on the seventh (Appendix~\ref{app:shared}).

Proposition~\ref{prop:ceiling} sets a hard ceiling on multi-model spending. Because every pair satisfies $G_{ij}\ge \gamma\min_k G_{kk}$ with $\gamma=0.75$ on older surveys and $0.86$ on new polls, no non-negative ensemble of these models can exceed 31 [27, 37] respondents on older surveys or 57 [50, 79] on new polls, a ceiling that also applies to future models with similar overlap. By Equation~\ref{eq:overlap}, a new model as strong as our best would add 20\% only if its error correlation with the best fell below 0.67, lower than in any GPT--Claude pair above. Without web search, all simulators likely draw from largely the same public record and miss the same information.

\paragraph{Agreement is not a safety check.} Top models cannot validate one another: the two strongest simulators sit almost three times closer to each other than to the truth (squared distance 0.011 between them, versus 0.028 and 0.031 from true shares), and when the best simulator misses by over five percentage points, the second-best errs in the same direction 93\% of the time \citep{wrongtogether} (Appendix~\ref{app:shared}).

\paragraph{Pooling with real respondents works out-of-sample.} Using Equation~\ref{eq:pool} in a real survey requires estimating $r$ from past surveys, which costs little accuracy. Estimating ensemble weights on six programmes and pooling with respondents from the seventh turns 5 real respondents into 29, 20 into 43, and 100 into 123, closely matching Proposition~\ref{prop:main}'s $n+25$ prediction; single simulators behave identically (Appendix~\ref{app:shared}).

\section{Real Answers Multiply a Simulator's Value}
\label{sec:real}

\begin{figure*}[t]
\centering
\includegraphics[width=0.95\textwidth]{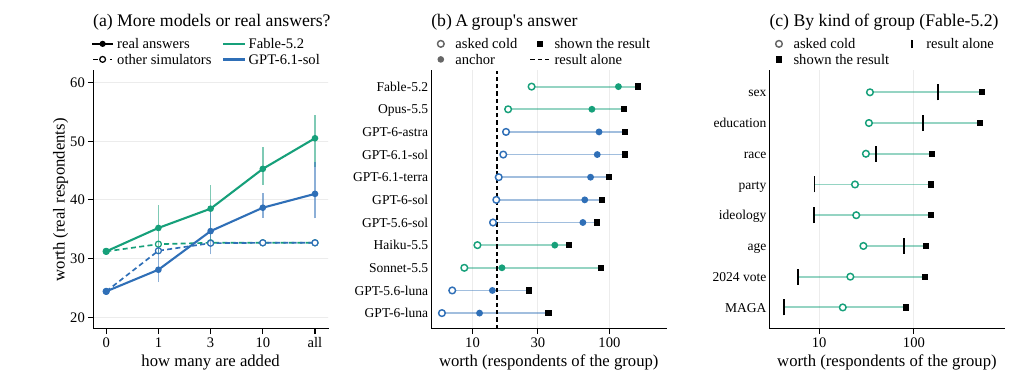}
\caption{Effect of real survey answers on 246 weekly poll questions. (a) Worth when adding 1, 3, 10, or all other simulators (in-sample non-negative weights; dashed) versus placing that many same-survey real results in the prompt (solid; mean and range over five splits; ``all'' = twelve models or half the survey, averaging 29 questions). (b) Subgroup worth across twenty groups (log scale): cold, anchored on the real overall result, and shown that overall result in the prompt (dashed line: overall result alone). (c) Breakdown of (b) for Claude-Fable-5.2.}
\label{fig:real}
\end{figure*}

Since extra models buy almost nothing, we test spending budget on real survey answers instead. On the 246 weekly poll questions (tested across the thirteen direct-output simulators; Figure~\ref{fig:real}a), cold Claude-Fable-5.2 is worth 31.2 respondents. Adding all twelve other simulators with optimal in-sample weights adds only 1.5 respondents (to 32.7). By contrast, placing real results from other questions in the same survey into its prompt raises worth to 38.5 with three questions, 45.3 with ten, and 50.5 with half the survey (means over five random splits). For GPT-6.1-sol (24 cold), combining it with all twelve other simulators (including Claude-Fable-5.2) reaches only the 13-model ceiling of 32.7, whereas showing GPT-6.1-sol alone half the survey reaches 41.0.

\subsection{The other questions of the same survey}

We randomly split each weekly report's questions into two halves, place one half's real results (29 questions on average) in the prompt, and query the other half. On the first split, all thirteen simulators gain 1.2 to 2.0 times over cold prompting, with every 95\% interval above one (Table~\ref{tab:ctx}); strong GPT models and Claude-Fable-5.2 gain most. Which questions are shown matters: across five random splits, Claude-Fable-5.2 ranges from 46 to 54 and GPT-6.1-sol from 37 to 47.

\paragraph{The survey matters more than the date.} Do shown answers help merely by updating post-cutoff world knowledge, or because they come from the same survey? On reports 2 through 5 (206 questions), showing 29 recent answers from same-week U.S.\ polls by other pollsters or from the prior week's report barely helps (Table~\ref{tab:ctl}): Claude-Fable-5.2 scores 30 cold, 31 with other pollsters, 33 with the prior week, and 54 with the same survey (GPT-6.1-sol: 24, 27, 27, and 49). Only when no shown item shares the target's response options do external recent answers account for up to half of top models' gain.

\paragraph{Question batteries help most.} Surveys often group related items into batteries, such as favourability ratings of several figures. For the 61 of 246 weekly questions where the prompt includes a \emph{sibling item} from the same battery, every simulator gains most, exceeding fivefold for the strongest models. On the 185 items without a sibling, four strong simulators still gain 28\% to 41\% (the other nine gain 2\% to 19\%; Appendix~\ref{app:context}). Other datasets behave similarly: Claude-Fable-5.2 rises from 58 to 79 and GPT-6.1-sol from 35 to 54 on 179 questions from eight other pollsters, and both gain roughly one fifth on 815 older items.

\paragraph{Prior waves of the same question.} For 44 weekly questions repeated within five weeks (median 12 days apart), the earlier wave alone is worth 286 respondents of the later wave, whereas cold simulators are worth 18 to 86 and add at most 1\% when pooled with it. Given the earlier wave in its prompt, Claude-Fable-5.2 shifts it by under one percentage point on average and reaches 390, roughly a third above the earlier wave alone, while other models' gains lie within noise (Appendix~\ref{app:wave}). Recent waves should therefore take precedence over fresh interviews or cold simulation.

\subsection{Groups and the overall result}

Surveys routinely report results by subgroup, such as women, younger voters, or party members. A simulator can be prompted for a subgroup directly, with its worth measured in respondents of that group. Across twenty groups, simply using the survey's overall result as the prediction for every group is worth 15 group respondents. Asked cold, the eleven GPT and Claude simulators range from 6 (GPT-6-luna) to 27 (Claude-Fable-5.2) group respondents. For broad groups whose views closely mirror the population average, such as men and women, the overall result alone is far stronger, worth 179 group respondents versus 34 for the best cold simulator.

Next, we place the question's real overall result in the prompt and ask for the twenty subgroup breakdowns (Figure~\ref{fig:real}b). Accuracy surges: Claude-Fable-5.2 jumps sixfold, from 27 to 163 group respondents; GPT-6.1-sol rises from 17 to 130; and even the weakest model, GPT-6-luna, climbs from 6 to 36. All eleven now outperform the overall result alone (15), and Claude-Fable-5.2 does so across every kind of group (Figure~\ref{fig:real}c).

Most of this surge comes from simple arithmetic anchoring. A simulator's errors for a subgroup and for the full population correlate strongly (0.68 to 0.89). Taking the real overall result and adding the simulator's cold predicted gap between the group and everyone cancels most of this shared error. For this anchor, Equation~\ref{eq:overlap} predicts a 1.6- to 4.7-fold gain over cold prompting, closely matching the measured anchor gain of 1.9- to 4.9-fold (within 21\% for every simulator). For Claude-Fable-5.2, anchoring alone raises worth from 27 to 116, and placing the overall result directly in the prompt pushes it further to 163.

\subsection{What the simulator contributes}

Real answers can also adjust a simulator's cold output outside the prompt. We compared prompting against both the best of several arithmetic corrections and a learned correction using gradient-boosted trees trained on roughly 200 other questions (Table~\ref{tab:ways}). For full-population answers, prompting wins clearly among top models: Claude-Fable-5.2 reaches 54 in-prompt versus 38 with arithmetic correction and 34 with the learned correction. For subgroups, the learned correction comes closer (149 versus 163), though it trains on real group results never shown to the simulator.

Most of this gain depends on the simulator itself, not on the real answers alone. If we remove the simulator and train the same gradient-boosted trees using only the question text and real survey data, worth drops sharply to 9 respondents for a population (from 34 with Claude-Fable-5.2) and 22 for a group (from 149). A standard hierarchical model that pools across groups without a simulator reaches only 16. Even when a small real sample is available for every group, the gap remains: combining 25 group respondents (sampled from published shares) with these non-simulator models via the small-area estimator of \citet{fayherriot} yields at most 46 respondents, compared with 186 when built on Claude-Fable-5.2 given the overall result.

Even with real answers in the prompt, extra simulators add little. When all thirteen models see the same half of the survey, a pair gains a median of 3\% over its stronger member, and all thirteen combined reach 61 (versus 54 for the best alone and 395 if counted independently). Nor can other models' predictions replace real data in the prompt: when we instead show each of six simulators the cold predictions of the two strongest models from the competing developer, none outperforms simply pooling those three models' cold predictions (Appendix~\ref{app:context}).

\subsection{Designing a survey under a fixed budget}

When a researcher can still allocate interviews across a new questionnaire, asking people directly beats cross-question conditioning: each respondent on a target question adds one respondent of worth, whereas 100 respondents on the other half add only 19 to Claude-Fable-5.2. Under a fixed budget of 50 interviews per question, asking 50 people every question and pooling with Claude-Fable-5.2 (Equation~\ref{eq:pool}) is worth 81 respondents overall. Asking 100 people half the questions (pooled on the asked half) and prompting with their answers on the unasked half reaches only 72, because overall worth follows the variance-weighted harmonic mean across halves. Spreading interviews across every item, as in split questionnaires \citep{splitq}, wins for all six tested simulators (63 to 81 versus 30 to 72; Appendix~\ref{app:context}).

When length or budget still forces omissions, selecting which items to field matters. Choosing ten questions per report via a greedy coverage rule maximizing $\sum_i \max_{j \in S} s_{ij}$ (TF-IDF cosine similarity $s_{ij}$) requires no survey answers and achieves a $1-1/e$ approximation guarantee for this coverage objective \citep{nemhauser1978}. On the remaining items, this rule cuts squared error by 10\% to 24\% relative to the mean of five random draws for five of six simulators, beating every individual draw for those five (Appendix~\ref{app:context}). Moreover, fielded anchor items need only small samples: replacing full-sample prompt results with 100-respondent samples keeps at least 79\% of the gain across all six simulators (50 versus 54 for Claude-Fable-5.2).

\subsection{A held-out replication}

Everything above was developed on polls up to 6 November. To verify our protocol out-of-sample, we held out the next two weekly reports (9 to 16 November 2026: 126 questions, including 53 not among the 246), froze the protocol, and ran it once on the eleven GPT and Claude simulators. All three findings hold. Individual simulators are worth 12 to 33 respondents; counted independently they sum to 198, yet combined they reach only 34, barely above the best alone (33). When shown the other half of the survey, they gain 1.5 to 2.5 times over cold prompting, with intervals above one for nine of the eleven (and 1.0 to 1.7 times on the 53 entirely new questions). When given the overall result, the eleven simulators gain 3.7 to 9.8 times over cold prompting, and every one outperforms the overall result alone (14).

\subsection{A six-point playbook for stretching a survey budget}

Sections~\ref{sec:ladder}--\ref{sec:real} translate into six rules for stretching a fieldwork budget:
\begin{enumerate}[leftmargin=*,itemsep=1pt,topsep=2pt]
    \item \textbf{Invest in one strong model, not an ensemble.} Fifteen extra models add fewer than two respondents, and model agreement does not confirm accuracy; focus compute on one frontier model.
    \item \textbf{Cover every question thinly from the start.} Spread interviews across all items and pool with a simulator (Equation~\ref{eq:pool}) rather than concentrating interviews on half the survey and simulating the rest.
    \item \textbf{When omitting items, field one per question group and cover topics broadly.} For similarly structured question sets (such as multi-figure ratings), fielding one item boosts the rest over fivefold; choose remaining anchor items for broad topical coverage (even 100-respondent samples keep most of the gain).
    \item \textbf{Solve subgroup estimation via the overall result.} Measure the full-population answer accurately and place it in the prompt rather than heavily oversampling subgroups (boosting the best model sixfold to 163 group respondents, or 186 with 25 real group respondents).
    \item \textbf{Deprioritize questions with recent waves.} Move items with a comparable recent measurement to the back of the queue (worth 286 alone, up to 390 in-prompt), reserving fresh interviews for new questions.
    \item \textbf{Protect human samples where simulators stay weak.} Keep full samples for populations underrepresented online, single high-stakes items, and analyses requiring cross-question correlations.
\end{enumerate}

\section{Related Work}
\label{sec:related}

\paragraph{Simulated respondents.} Persona-prompted LLMs were proposed as ``silicon samples'' \citep{argyle}, but evaluations find uneven, prompt-sensitive errors \citep{opinionqa,globalqa,simbench,questioning,meister}. Scaling \citep{ziems} and survey fine-tuning \citep{subpop,cao,kimlee} can improve fidelity, and LLMs can predict aggregate experimental effects \citep{hewitt}, yet personas explain little individual variation \citep{personaeffect,itemmean,funhouse}.

\paragraph{How many people is a model worth?} \citet{howmany} measure simulation fidelity through confidence-set coverage, and \citet{gaohanliang} via the task-specific training data needed to match an LLM. \citet{sizeweight} learn synthetic sample sizes and weights that satisfy task-marginal coverage. Prediction-powered inference and related methods use human data to correct model-based estimates for valid inference \citep{ppi,ppipp,mixedsubjects,rectify,weizhu,tanzrnic}. Our $r$ is closest to \citet{howmany}: applying their asymptotic lower-bound expression ($\alpha=0.05$) to individual option shares yields roughly half of $r$ on older surveys (median ratio 0.57; Spearman $\rho=0.91$), whereas the 95\% rather than 97.5\% error quantile gives 1.02 (Appendix~\ref{app:shared}). Defining $r$ via MSE makes it additive with independent human respondents under optimal weighting and extends it to ensembles (Proposition~\ref{prop:main}), connecting to effective-sample-size interpretations \citep{priorESS,meng,powerprior} and correlated forecast combination \citep{batesgranger,clemenwinkler}.

\paragraph{Shared errors.} LLMs make correlated errors \citep{kimcorr,goelalike}, and algorithmic monocultures can homogenize decisions \citep{kleinbergmono,sameperson}. For survey simulators, \citet{wrongtogether} shows across six models that shared errors can be hidden by model agreement and cancel when they affect two groups equally. We translate this overlap into equivalent respondents, capping non-negative ensembles and placing models, prior waves, and human responses on one budgeting scale.

\paragraph{Using real answers.} Prior work uses real data to correct model-based estimates \citep{ppi,rectify,pas,multippi}, allocate interviews \citep{yeyoga}, borrow strength for small areas \citep{fayherriot,mrp}, fine-tune models \citep{subpop,cao,kimlee}, or condition on past user opinions and interview transcripts \citep{hwang2023,parkagents}. Closely related, \citet{syndigits} calibrate unseen questions using human answers to reference items, including aggregate marginal distributions. We quantify same-survey answers in respondent units and compare their marginal gain against additional simulators. \citet{subpop} report limited gains from few-shot examples drawn from survey training data; in our setting, frontier models gain substantially when given real answers from the same survey.

\section{Limitations and Conclusion}
\label{sec:limits}

These guidelines carry clear boundaries. First, we score marginal shares per item; real respondents also preserve cross-question correlations that simulators can distort \citep{bisbee}. Second, worth varies fivefold or more across survey programmes, so average worth cannot guarantee accuracy on any single high-stakes question. Third, we benchmark against simple random draws from published shares: for an estimate with design effect $d$, a complex survey of $n$ respondents has an effective sample size of $n/d$ \citep{kish1965}, and published shares may also reflect non-sampling house or mode effects that our sampling-noise adjustment does not remove.

Scope is likewise bounded: top models come from two developers, open models stop at 27B parameters, and search-enabled or survey-tuned models may escape our ceiling. Most new polls cover politics around one U.S.\ election, main subgroup results use one pollster, unit-level models such as MRP require unpublished microdata, and splitting fielded surveys shares sample and house effects between prompt and target.

A survey simulator is worth roughly a small poll, and gains come from a better model, not more models: shared errors leave sixteen models barely above one. Yet that same overlap across related questions makes real answers powerful in the prompt, lifting a strong simulator by roughly half on full populations and sixfold on subgroups. Stretching a survey budget is therefore not about replacing respondents, but about placing scarce interviews where they multiply a model's value most.

\section*{Impact Statement}

This paper evaluates language-model survey simulators and offers guidelines for stretching survey budgets alongside real respondents. Our core caution is that polling many models cannot substitute for real people, and simulations are most reliable when grounded in same-survey data. We use only published aggregate results and group-level shares from the public General Social Survey, releasing our code, prompts, model outputs, and data-rebuilding scripts.

\paragraph{AI Disclosure Statement.} Beyond studying language models as simulators (Section~\ref{sec:setup}, Appendix~\ref{app:details}), we used the AI coding agent Claude Code for code revision, experimental setup, and grammar. All AI-assisted outputs were reviewed and verified by the authors.

\bibliography{references}
\bibliographystyle{icml2027}

\newpage
\appendix
\onecolumn

\section{Proofs}
\label{app:proof}

Throughout, we write $W=\mathbf{1}^\top w$ for the total weight assigned to the sources and $\varepsilon=\bar x-\theta$ for the sampling error of $n$ real respondents, which satisfies $\mathbb{E}[\varepsilon\mid\theta]=0$ and $\mathbb{E}\,\|\varepsilon\|^2=V/n$.

\paragraph{Proposition~\ref{prop:main} (Optimal pooling of respondents and sources).}
The error of the pooled estimate $(n\bar x+\sum_j w_jc_j)/(n+W)$ is $(n\varepsilon+\sum_j w_j e_j)/(n+W)$. Because the real respondents are sampled independently of the sources, $\varepsilon$ has conditional mean zero given $\theta$ and the sources, so all cross terms between $\varepsilon$ and $e_j$ vanish. Taking expectations yields the mean squared error
\[
L(w)
=\frac{n^2\,\mathbb{E}\,\|\varepsilon\|^2+w^\top G\,w}{(n+W)^2}
=\frac{nV+w^\top G\,w}{(n+W)^2}.
\]
Because $G$ is positive definite, let $s=\mathbf{1}^\top G^{-1}\mathbf{1}>0$. By the Cauchy--Schwarz inequality,
\[
W^2=(\mathbf{1}^\top w)^2=\big((G^{-1/2}\mathbf{1})^\top(G^{1/2}w)\big)^2\le (\mathbf{1}^\top G^{-1}\mathbf{1})(w^\top G\,w)=s\,w^\top G\,w,
\]
with equality if and only if $w\propto G^{-1}\mathbf{1}$. Substituting $w^\top G\,w\ge W^2/s$ into $L(w)$ gives, for every $w$ with $n+W>0$,
\[
L(w)-\frac{V}{n+Vs}
\ge \frac{nV+W^2/s}{(n+W)^2}-\frac{V}{n+Vs}
=\frac{n(Vs-W)^2}{s(n+W)^2(n+Vs)}\ge 0.
\]
Equality holds at $w^\star=V\,G^{-1}\mathbf{1}$ (for which $W=Vs$ and $w^{\star\top}G\,w^\star=V^2s$), uniquely when $n>0$ and up to positive rescaling when $n=0$. Hence $w^\star=V\,G^{-1}\mathbf{1}$ is globally optimal regardless of $n$, and the minimum mean squared error is
\[
L(w^\star)=\frac{V}{n+Vs},
\]
matching a random sample of $n+r_S$ respondents with $r_S=Vs=V\,\mathbf{1}^\top G^{-1}\mathbf{1}$.

\paragraph{Sensitivity to misspecified weight (one source).}
Suppose a single source with true worth $r$ is given a misspecified weight $w=\kappa r$ ($\kappa>0$). Writing $x=n/r$ for the ratio of real respondents to the source's worth, the pooled MSE divided by the respondents-only MSE ($V/n$) is
\[
f(x)=\frac{nV+(\kappa r)^2(V/r)}{(n+\kappa r)^2}\cdot\frac{n}{V}=\frac{x^2+\kappa^2 x}{(x+\kappa)^2}.
\]
Note that $f(x)<1$ if and only if $\kappa(\kappa-2)x<\kappa^2$. Thus, for any $\kappa\in(0,2]$, $f(x)<1$ strictly for every $x>0$ (pooling always helps). For $\kappa>2$, maximizing $f(x)$ over $x>0$ gives a unique maximum at $x=\kappa^2/(\kappa-2)$, where $f=\kappa^2/(4(\kappa-1))$. At $\kappa=3$, this worst-case penalty is $9/8=1.125$ (at most 12.5\% higher MSE).

\paragraph{Nested errors and non-negative weights.}
Order the sources so that $G_{11}\le G_{jj}$ for all $j$, and suppose errors are nested: $G_{ij}=\min(G_{ii},G_{jj})$. Then $G_{j1}=G_{11}$ for all $j$, so $w^\star=(V/G_{11},0,\dots,0)=(r_1,0,\dots,0)$ directly satisfies $G\,w^\star=V\mathbf{1}$, giving $r_S=r_1$.

For two sources in general, inverting the $2\times 2$ matrix $G$ yields $w^\star\propto(G_{22}-G_{12},\,G_{11}-G_{12})$, so the weaker source ($j=2$) receives positive weight if and only if $G_{12}<G_{11}$ \citep{clemenwinkler}. If both sources have equal worth $r$ and error correlation $\rho$, then $G_{11}=G_{22}=V/r$ and $G_{12}=\rho V/r$, giving $r_S=2r/(1+\rho)$.

In our empirical ensembles, whenever unconstrained $w^\star$ contains negative weights, we report the best non-negative combination: the exact solution of the convex problem $\min_{w\ge0}\tfrac12w^\top G\,w-V\mathbf{1}^\top w$, computed by non-negative least squares after a Cholesky factorization of $G$ (positive definite in every case we report). By KKT complementarity, the solution satisfies $w^\top G\,w=V\mathbf{1}^\top w$ and minimizes $p^\top G\,p$ over the simplex $p=w/(\mathbf{1}^\top w)$, so it is simultaneously optimal among all $w\ge 0$ for every $n\ge 0$ and achieves worth $n+\mathbf{1}^\top w$. Allowing negative weights changes the combined worth of the sixteen simulators only slightly, from 24.8 to 25.2.

\paragraph{Proposition~\ref{prop:ceiling} (Overlap ceiling).}
Any non-negative combination of sources corresponds to normalized weights $p\ge 0$ with $\mathbf{1}^\top p=1$ and achieves worth $V/(p^\top G\,p)$. If $G_{ij}\ge\gamma\,\min_k G_{kk}$ for all $i,j$ with $\gamma\in(0,1]$ (note that $i=j$ holds automatically whenever $\gamma\le 1$), then for every such $p$,
\[
p^\top G\,p
=\sum_{i,j} p_i p_j G_{ij}
\ge \gamma\,\min_k G_{kk}\sum_{i,j} p_i p_j
=\gamma\,\min_k G_{kk}.
\]
Consequently, no non-negative combination can be worth more than $V/(\gamma\min_k G_{kk})=r_{\max}/\gamma$ respondents.

\paragraph{Equation~\ref{eq:overlap} (Second model versus real anchor).}
Consider two errors $e,e'$ with equal mean squared norm $G$ and correlation $\rho=\mathbb{E}\langle e,e'\rangle/G$.
\begin{itemize}[leftmargin=*,itemsep=2pt,topsep=2pt]
\item \emph{Averaging two models} has error $(e+e')/2$ and MSE $\frac{1}{4}(G+G+2\rho G)=\frac{1+\rho}{2}G$, multiplying worth by $2/(1+\rho)$.
\item \emph{Anchoring on a real answer} via $\theta'+(c-c')$ has error $e-e'$ and MSE $G+G-2\rho G=2(1-\rho)G$, multiplying worth by $1/(2(1-\rho))$.
\end{itemize}
More generally (without assuming equal error sizes), the anchor's MSE is $\mathbb{E}\|e\|^2+\mathbb{E}\|e'\|^2-2\,\mathbb{E}\langle e,e'\rangle$, which improves on the unanchored prediction ($\mathbb{E}\|e\|^2$) if and only if $\mathbb{E}\langle e,e'\rangle>\tfrac12\mathbb{E}\|e'\|^2$, or for equal-sized errors, if and only if $\rho>1/2$.

\section{Data and Protocol}
\label{app:details}

\paragraph{Older surveys.} Our full pool contains 3,041 items: 41 from the General Social Survey and 500 each from the OpinionQA, European Social Survey, Afrobarometer, ISSP, Latinobar\'ometro, and GlobalOpinionQA subsets of SimBench \citep{simbench,opinionqa,globalqa}, where an item is one question asked of one national population. We grouped each programme's items into batches of twenty, interleaved programmes round-robin, and evaluated every simulator on the first fifty batches (981 items, as the GSS contributes only 41). Because GPT-6-astra failed to return one batch of twenty GSS items, all main comparisons use the remaining 961 common items.

\paragraph{New polls and weekly reports.} The 250 new-poll items are single-choice questions with complete published response distributions whose fieldwork began between 1 October and 6 November 2026, drawn from nine pollsters: The Economist/YouGov (60 questions), YouGov UK (33), AP-NORC (32), Reuters/Ipsos (30), Essential Research (28), YouGov EuroTrack (24), NPR/PBS News/Marist (22), Angus Reid (16), and Gallup (5). By country, 149 come from the United States, 37 from the United Kingdom, 28 from Australia, 16 from Canada, and four each from France, Germany, Denmark, Spain, and Italy. Party-identification splits are published for 113 of these questions.

Separately, an automated parser extracted all five weekly Economist/YouGov reports from the same period (1,381 to 3,166 registered voters per report), retaining the 246 distinct questions that report complete distributions for both all respondents and all twenty subgroups. Recurring tracking questions are notably easier: on the 38 of 246 questions asked in more than one report, the best simulator is worth 75 respondents cold, versus 28 on the 208 questions asked only once. Crucially, the in-prompt real-answer gains of Section~\ref{sec:real} are just as strong on the one-off questions (1.2 to 2.1 times). Note that Appendix~\ref{app:wave} counts 44 recurring questions because that analysis requires only the party table (not all twenty groups) to be complete. Source URLs and file-version checks are recorded in \texttt{data/download\_list.tsv} and \texttt{data/sources.json} in our code repository.

\paragraph{Prompting and execution.} All GPT and Claude models were invoked via their developers' non-interactive CLI clients with tools disabled, medium reasoning effort, and a single enforced JSON output schema. Mistral-Small-24B and Qwen3.8-27B were served locally in 4-bit quantization with the same schema enforced. Unless stated otherwise, each API call contained twenty items prefixed by the instruction:
\begin{quote}\small\itshape
``You are estimating the results of real opinion surveys. For each item below, a group of respondents is described and a question is given with its answer options. Estimate the percentage of that group who chose each option in the actual survey (percentages over the listed options, summing to 100). Answer from your own knowledge only: do not run commands, read files or search.''
\end{quote}
Any batch failing schema validation after three attempts was re-queried in smaller batches, down to one question per call. Primary runs took place on 10--11 November 2026, with remaining table entries completed on 24--25 November 2026. When queried for their training cutoffs, GPT models report dates between June 2024 and August 2025, Claude models report June 2026, and all five open models were released prior to October 2026.

\paragraph{Handling the three smallest open models.} OLMo-3-7B, Qwen3.5-9B, and Gemma-4-12B struggle to output well-calibrated aggregate percentages directly. Instead, we address them in the second person as a member of the target population, list lettered options, and append ``Answer with a single letter.'' We extract their next-token probability distribution $p$ over option letters, averaged across both forward and reversed option orders to reduce position bias. Because raw token probabilities are overconfident and under-predict non-substantive options (worth only 1.8, 3.9, and 1.3 respondents raw, below the 4.4 uniform baseline), we apply a two-parameter calibration:
\[
\tilde p \propto \mathrm{softmax}\!\left(\frac{\log p}{T} - \delta\,\mathbf{1}_{\text{non-answer}}\right),
\]
where $\mathbf{1}_{\text{non-answer}}$ flags options such as ``don't know'' or ``refused.'' Both $T$ and $\delta$ are fitted out-of-sample on the other six programmes (or on all older items when evaluating new polls). For the party-group experiments in Appendix~\ref{app:groups}, these three models instead output percentages directly.

\paragraph{Protocol robustness checks.} To verify that our conclusions do not depend on prompting or scoring choices, we ran six sensitivity checks:
\begin{itemize}[leftmargin=*,itemsep=2pt,topsep=2pt]
\item \textbf{Stated percentages vs.\ next-token probabilities:} Asking Qwen3.5-9B for stated percentages yields 5.7 respondents raw and 6.5 rescaled, close to 7.0 via next-token probabilities (error correlation 0.84 between methods).
\item \textbf{Panel of individual personas:} Averaging responses from sixteen simulated personas answering as individuals yields 7.5 respondents for Qwen3.5-9B (vs.\ 7.0 direct), 22.0 for Claude-Fable-5.2 (vs.\ 23.2), and 16.0 for GPT-6.1-sol (vs.\ 20.3), showing no consistent gain over direct share estimation.
\item \textbf{Rescaling larger models:} Applying the same $(T,\delta)$ rescaling to stated percentages of GPT and Claude models changes their worth only slightly (10.2 to 21.7 vs.\ 9.6 to 23.2 raw). Under rescaling, error overlap is even tighter (pooled overlap ratio 0.94 [0.91, 0.96]), and all sixteen combined reach 22.7 versus 21.7 for the best alone.
\item \textbf{Repeat runs and rewording:} Across all 961 items, two independent runs of identical batches have errors that correlate 0.94 (GPT-6.1-sol), 0.96 (GPT-6-astra), 0.92 (Claude-Opus-5.5), and 0.95 (Claude-Fable-5.2), with their average gaining 1\% to 3\% over the better run. An alternative prompt wording correlates 0.94 (Claude-Fable-5.2), 0.93 (GPT-6.1-sol), 0.86 (GPT-5.6-luna), and 0.82 (Claude-Haiku-5.5) with the original, gaining 2\% to 9\% when averaged.
\item \textbf{Single-item calls vs.\ batches of twenty:} On all 961 items, asking one question per call yields 26.5 (Claude-Fable-5.2), 20.5 (Claude-Opus-5.5), 20.1 (GPT-6-astra), and 19.4 (GPT-6.1-sol), compared with 23.2, 18.7, 20.7, and 20.3 in batches of twenty. Error correlations remain 0.80 to 0.94, and combining all four reaches only 27.0.
\item \textbf{Alternative loss (KL divergence):} Replacing squared error with KL divergence preserves simulator rankings; two-model mixtures reduce divergence by a median of 7.6\% over their better member, while the unweighted mean of all sixteen is 20\% worse than the best model alone.
\end{itemize}

\section{What Simulators Are Worth}
\label{app:shared}

Tables~\ref{tab:ladder} and~\ref{tab:prog} report individual worth across all sixteen simulators, non-model baselines, survey programmes, and poll subsets. Below we summarize six additional analyses of single-model and ensemble performance.

\begin{table}[!htb]
\centering\small
\setlength{\tabcolsep}{3.2pt}
\begin{tabular}{lcccccccc}
\toprule
 & \multicolumn{2}{c}{Population (respondents)} & & \multicolumn{3}{c}{Party groups, new polls} & \multicolumn{2}{c}{20 groups, GSS} \\
\cmidrule(lr){2-3}\cmidrule(lr){5-7}\cmidrule(lr){8-9}
Source & 961 older questions & 250 new polls & & cold & +difference & shown & cold & +difference \\
\midrule
\emph{Uniform answer} & 4.4 & 5.4 & & & & & & \\
\emph{Same-scale answers} & 8.1 & 13.4 {\scriptsize($n$=101)} & & & & & & \\
\emph{Another population} & 9.8 & 16.9 {\scriptsize($n$=24)} & & & & & & \\
\emph{The overall result alone} & & & & \multicolumn{3}{c}{11 {\scriptsize[9, 13]}} & \multicolumn{2}{c}{36 {\scriptsize[30, 45]}} \\
\midrule
OLMo-3-7B & 6.2 {\scriptsize[6, 7]} & 6.8 {\scriptsize[6, 9]} & & 3$^\dagger$ & 6$^\dagger$ & 3$^\dagger$ & 6 & 34 \\
Mistral-Small-24B & 6.9 {\scriptsize[6, 8]} & 9.1 {\scriptsize[8, 11]} & & 6 & 13 & 14 & 8 & 28 \\
Qwen3.5-9B & 7.0 {\scriptsize[6, 8]} & 9.3 {\scriptsize[8, 11]} & & 4$^\dagger$ & 7$^\dagger$ & 6$^\dagger$ & 5 & 40 \\
Gemma-4-12B & 7.1 {\scriptsize[6, 8]} & 8.0 {\scriptsize[7, 10]} & & 5$^\dagger$ & 9$^\dagger$ & 9$^\dagger$ & 8 & 28 \\
Qwen3.8-27B & 7.5 {\scriptsize[7, 8]} & 10.6 {\scriptsize[10, 12]} & & 6 & 15 & 20 & 11 & 44 \\
GPT-6-luna & 9.6 {\scriptsize[9, 11]} & 14.4 {\scriptsize[12, 17]} & & 8 & 17 & 36 & 23 & 103 \\
GPT-5.6-luna & 9.8 {\scriptsize[9, 11]} & 14.4 {\scriptsize[12, 17]} & & 8 & 16 & 24 & 21 & 108 \\
Claude-Haiku-5.5 & 10.5 {\scriptsize[10, 11]} & 16.9 {\scriptsize[15, 20]} & & 13 & 41 & 55 & 22 & 93 \\
Claude-Sonnet-5.5 & 12.2 {\scriptsize[11, 14]} & 19.9 {\scriptsize[16, 25]} & & 13 & 30 & 26 & 30 & 151 \\
GPT-5.6-sol & 14.2 {\scriptsize[13, 16]} & 21.6 {\scriptsize[18, 26]} & & 17 & 57 & 79 & 33 & 159 \\
GPT-6-sol & 15.6 {\scriptsize[14, 17]} & 26.2 {\scriptsize[21, 32]} & & 20 & 62 & 83 & 45 & 170 \\
GPT-6.1-terra & 17.2 {\scriptsize[16, 19]} & 25.8 {\scriptsize[20, 32]} & & 19 & 62 & 85 & 61 & 212 \\
Claude-Opus-5.5 & 18.7 {\scriptsize[17, 21]} & 32.4 {\scriptsize[28, 38]} & & 25 & 73 & 96 & 110 & 352 \\
GPT-6.1-sol & 20.3 {\scriptsize[18, 23]} & 32.2 {\scriptsize[25, 40]} & & 22 & 60 & 101 & 61 & 231 \\
GPT-6-astra & 20.7 {\scriptsize[18, 23]} & 34.6 {\scriptsize[27, 43]} & & 24 & 66 & 102 & 60 & 146 \\
Claude-Fable-5.2 & 23.2 {\scriptsize[21, 26]} & 49.1 {\scriptsize[40, 62]} & & 35 & 99 & 128 & 169 & 491 \\
\bottomrule
\end{tabular}
\caption{Worth of each source in respondents (with 95\% bootstrap intervals). Left: full-population answers, compared against three non-model baselines (\emph{Uniform answer}: equal shares; \emph{Same-scale answers}: average real distribution on other questions sharing the same response options; \emph{Another population}: real answer to the same question in another country/population, tempered by a power of 0.75, vs.\ 8.5 and 16.0 untempered; $n$ counts new-poll items where each baseline exists). Right: subgroup answers on two additional datasets (Appendix~\ref{app:groups}), in group respondents, asked cold, anchored (+difference), or given the overall result in the prompt. For the three smallest open models, we use rescaled next-token probabilities except on party groups, where they state percentages ($^\dagger$).}
\label{tab:ladder}
\end{table}

\begin{table}[!htb]
\centering\small
\setlength{\tabcolsep}{3.6pt}
\begin{tabular}{lccccccccccc}
\toprule
 & \multicolumn{7}{c}{Survey programme} & all, re- & \multicolumn{3}{c}{New polls} \\
\cmidrule(lr){2-8}\cmidrule(lr){10-12}
Source & GSS & OpinionQA & ESS & Afrobar. & ISSP & Latinobar. & GlobalOQA & scaled & U.S. & other & no tracker \\
\midrule
\emph{Items} & 21 & 160 & 160 & 160 & 160 & 160 & 140 & & 149 & 101 & 234 \\
\emph{Uniform answer} & 7.3 & 3.0 & 7.3 & 3.5 & 5.1 & 5.5 & 3.5 & 4.4 & 4.6 & 7.2 & 5.5 \\
\midrule
OLMo-3-7B & 8 & 8 & 9 & 5 & 7 & 5 & 5 & 6.2 & 6 & 9 & 7 \\
Mistral-Small-24B & 12 & 7 & 11 & 5 & 8 & 8 & 5 & 7.2 & 9 & 10 & 9 \\
Qwen3.5-9B & 11 & 7 & 10 & 6 & 8 & 6 & 6 & 7.0 & 9 & 10 & 10 \\
Gemma-4-12B & 13 & 8 & 10 & 6 & 8 & 6 & 6 & 7.1 & 8 & 8 & 8 \\
Qwen3.8-27B & 19 & 12 & 11 & 5 & 9 & 7 & 5 & 7.9 & 11 & 10 & 11 \\
GPT-6-luna & 27 & 18 & 16 & 7 & 11 & 7 & 6 & 10.2 & 16 & 12 & 14 \\
GPT-5.6-luna & 21 & 17 & 13 & 7 & 11 & 7 & 8 & 10.2 & 16 & 12 & 14 \\
Claude-Haiku-5.5 & 17 & 14 & 17 & 8 & 10 & 10 & 7 & 10.4 & 18 & 16 & 17 \\
Claude-Sonnet-5.5 & 34 & 20 & 23 & 9 & 11 & 11 & 8 & 12.2 & 25 & 15 & 19 \\
GPT-5.6-sol & 32 & 27 & 24 & 10 & 15 & 12 & 9 & 14.4 & 21 & 22 & 21 \\
GPT-6-sol & 62 & 28 & 26 & 11 & 16 & 15 & 10 & 15.5 & 30 & 22 & 26 \\
GPT-6.1-terra & 56 & 29 & 31 & 12 & 18 & 18 & 10 & 16.6 & 30 & 21 & 25 \\
Claude-Opus-5.5 & 91 & 40 & 35 & 13 & 19 & 17 & 11 & 17.6 & 36 & 28 & 31 \\
GPT-6.1-sol & 59 & 30 & 36 & 16 & 24 & 17 & 12 & 19.2 & 42 & 25 & 31 \\
GPT-6-astra & 56 & 38 & 36 & 16 & 25 & 17 & 12 & 19.4 & 44 & 27 & 34 \\
Claude-Fable-5.2 & 345 & 79 & 45 & 17 & 27 & 17 & 13 & 21.7 & 54 & 44 & 48 \\
\bottomrule
\end{tabular}
\caption{Simulator worth broken down by survey programme and subsets of the 250 new polls. ``All, rescaled'': worth across all 961 older questions after two-parameter out-of-sample rescaling. ``No tracker'': the 234 new-poll questions excluding regularly repeated tracking items such as presidential approval.}
\label{tab:prog}
\end{table}

\paragraph{How well the nested-error model fits.} Predicting the off-diagonal covariances $G_{ij}$ on one half of the questions from the other half (200 random splits), the zero-parameter nested model ($G_{ij}=\min(G_{ii},G_{jj})$) achieves an RMSE of 0.165 in correlation units, nearly matching a 16-parameter one-factor model (0.150), against a sampling-noise floor of 0.071. Nesting is nevertheless approximate: the weaker model's optimal weight has a 95\% interval excluding zero in 119 of 120 pairs, yet adds at most 17\% in worth. Within individual survey programmes, pooled overlap ratios range from 0.70 to 0.91, and all sixteen models together beat the best single model by just 4\% to 13\%. Directional agreement in errors is also striking: whenever the best simulator misses a share by more than five percentage points, the second-best model errs in the same direction 93\% of the time.

\paragraph{Alternative ensemble combiners.} Fitting each combiner on six programmes and evaluating on the held-out seventh gives almost identical ceilings (compared against 23.2 for the best single simulator):
\begin{itemize}[leftmargin=*,itemsep=1pt,topsep=1pt]
\item \emph{Convex pooling} of the top two, top four, or all sixteen simulators: 24.3 to 24.4 respondents.
\item \emph{Linear stacking} with unconstrained coefficients and an intercept: 23.9 to 24.4 respondents.
\item \emph{Log-linear pooling} with a fitted exponent: 16.2 to 22.4 respondents.
\item \emph{Disagreement-dependent weights} conditioned on the gap between Claude-Fable-5.2 and GPT-6-astra: identical to constant weights.
\end{itemize}

\paragraph{Question-level dispersion.} Across individual questions, the effective respondent count whose expected squared error matches Claude-Fable-5.2's realized error (mean worth 23.2) has quartiles 22, 59, and 179. The worst tenth of questions contributes 51\% of total squared error. Treating Claude-Fable-5.2's predictions as if they came from 23.2 random respondents produces nominal 95\% intervals covering 96\% of published option shares on average, though this reflects aggregate calibration rather than a per-question guarantee.

\paragraph{Transferring estimated $r$ across surveys.} For every pair of simulator (16) and target dataset (8: seven programmes plus new polls), we estimated $r$ out-of-sample (from the other six programmes, or from older surveys for new polls) and pooled the simulator with $n$ real respondents via Equation~\ref{eq:pool}. The transferred $r$ has a median ratio of 0.84 to the true $r$ (range 0.07 to 2.03). Despite this spread, pooling beats real respondents alone in all 128 cases at $n=5$, 20, and 100, retaining a median of 95\%, 90\%, and 86\% of the gain achievable with the oracle $r$.

\paragraph{Human-preference annotators.} Beyond opinion polls, we evaluated twelve simulators on Community Alignment \citep{commalign} (130 prompts, four candidate responses each, $\ge 26$ human annotators per prompt, with annotator sampling variance subtracted). Individual simulators are worth 5.0 to 14.3 annotators, with a pooled overlap ratio of 0.85 [0.80, 0.90]. Counted independently they sum to 110 annotators, yet combined they reach only 15.8 (versus 14.3 for the best alone).

\paragraph{Comparison with coverage-based effective sample size.}
Using the asymptotic expressions in Theorem~4.2 of \citet{howmany}, we compute the plug-in comparison
\[
    \kappa_q = \frac{z_{\alpha/2}^{2}}{Q_q(\Delta)},
    \qquad \Delta = \frac{(c-h)^2}{c(1-c)},
\]
where $c$ and $h$ are the simulated and published shares of an individual answer option, $z_{\alpha/2}=\Phi^{-1}(1-\alpha/2)$ is the standard normal critical value, and $Q_q$ is the empirical $q$-quantile. We set $\alpha=0.05$ and evaluate each simulator separately. Options are pooled with equal weight, so questions with more options contribute more to the quantile; we use linear interpolation between adjacent ordered discrepancies. For numerical stability, the implementation replaces $c(1-c)$ by $\max\{c(1-c),10^{-12}\}$. We treat each option as a binary outcome and use published shares as proxies for population proportions, rather than running their finite-sample procedure.

Across sixteen simulators on the 961 older questions, the median ratio $\kappa_q/r$ is 0.57, 1.02, and 1.93 for $q=0.975$, $0.95$, and $0.90$, respectively, with corresponding Spearman rank correlations of 0.91, 0.93, and 0.96. The first and third quantiles match those in their asymptotic lower- and upper-bound expressions; the middle is a sensitivity comparison, not an estimate of their finite-sample oracle. The absolute scale therefore depends on the quantile choice, while model rankings remain similar. This plug-in comparison does not establish a coverage guarantee for our predictions.

\FloatBarrier
\section{Real Answers}
\label{app:context}

\paragraph{In-prompt real answers from the same survey.} Within each of the five weekly reports, questions are randomly split into two halves. In the \emph{shown} condition, the prompt names the survey (population and month), lists every question from the other half with its options and published percentages under ``Actual results of other questions in the same survey,'' and asks for up to twenty target questions from the held-out half. The \emph{cold} condition uses the exact same prompt without that list. Published-share sampling noise is subtracted, and questions within a report sharing a common stem are resampled together in the bootstrap.

Key robustness details: (i) querying one target question per call reproduces the twenty-question batch ratio (1.78 vs.\ 1.74 for Claude-Fable-5.2; 1.92 vs.\ 1.91 for GPT-6.1-sol); (ii) in the small-sample variant, each shown result is replaced by a synthetic sample of 100 respondents drawn from its published shares (with the sample size noted in the prompt), and the six tested simulators keep 79\% or more of their gain; and (iii) on the 185 questions without any sibling item in the prompt, the shown-to-cold ratio of the four simulators that gain most remains well above one: 1.28 [1.11, 1.46] for Claude-Fable-5.2, 1.41 [1.14, 1.77] for GPT-6-astra, 1.41 [1.14, 1.76] for GPT-6.1-sol, and 1.32 [1.11, 1.61] for GPT-6.1-terra, while for the other nine it is 1.02 to 1.19. Full results appear in Tables~\ref{tab:ctx} and~\ref{tab:ctl}.

\begin{table}[!htb]
\centering\small
\setlength{\tabcolsep}{4pt}
\begin{tabular}{lccccccc}
\toprule
 & \multicolumn{3}{c}{All 246 questions} & \multicolumn{2}{c}{Sibling item shown (61)} & \multicolumn{2}{c}{No shared options (140)} \\
\cmidrule(lr){2-4}\cmidrule(lr){5-6}\cmidrule(lr){7-8}
Simulator & cold & shown & ratio & cold & shown & cold & shown \\
\midrule
Mistral-Small-24B & 9 & 11 & 1.19 {\scriptsize[1.08, 1.35]} & 15 & 24 & 8 & 9 \\
Qwen3.8-27B & 10 & 15 & 1.46 {\scriptsize[1.16, 1.88]} & 9 & 29 & 10 & 12 \\
Claude-Haiku-5.5 & 13 & 18 & 1.36 {\scriptsize[1.08, 1.72]} & 13 & 32 & 12 & 14 \\
GPT-5.6-luna & 15 & 19 & 1.27 {\scriptsize[1.02, 1.63]} & 13 & 36 & 14 & 14 \\
GPT-6-luna & 15 & 21 & 1.34 {\scriptsize[1.02, 1.76]} & 12 & 40 & 15 & 16 \\
Claude-Sonnet-5.5 & 18 & 25 & 1.37 {\scriptsize[1.05, 1.84]} & 16 & 43 & 18 & 21 \\
GPT-5.6-sol & 19 & 29 & 1.53 {\scriptsize[1.06, 2.24]} & 14 & 65 & 20 & 22 \\
GPT-6-sol & 21 & 32 & 1.49 {\scriptsize[1.01, 2.28]} & 15 & 68 & 23 & 25 \\
GPT-6.1-terra & 22 & 37 & 1.72 {\scriptsize[1.22, 2.57]} & 16 & 66 & 22 & 31 \\
GPT-6.1-sol & 24 & 47 & 1.91 {\scriptsize[1.26, 2.94]} & 16 & 65 & 27 & 42 \\
GPT-6-astra & 26 & 52 & 1.99 {\scriptsize[1.22, 3.08]} & 17 & 89 & 29 & 46 \\
Claude-Opus-5.5 & 28 & 36 & 1.32 {\scriptsize[1.04, 1.76]} & 26 & 76 & 26 & 29 \\
Claude-Fable-5.2 & 31 & 54 & 1.74 {\scriptsize[1.19, 2.69]} & 23 & 125 & 33 & 43 \\
\bottomrule
\end{tabular}
\caption{Full-population worth (in respondents) when real results from the other half of the same weekly report (29 questions on average) are shown in the prompt. \emph{Ratio}: shown divided by cold, with 95\% interval. \emph{Sibling item shown}: another item from the same question battery appears in the prompt. \emph{No shared options}: no shown question shares the target's exact response options. Values reflect one random split; across five splits, Claude-Fable-5.2 averages 50.5 and GPT-6.1-sol averages 41.0 (Figure~\ref{fig:real}a).}
\label{tab:ctx}
\end{table}

\begin{table}[!htb]
\centering\small
\setlength{\tabcolsep}{4.5pt}
\begin{tabular}{lcccccccc}
\toprule
 & \multicolumn{4}{c}{All 206 questions} & \multicolumn{4}{c}{No shared options (107)} \\
\cmidrule(lr){2-5}\cmidrule(lr){6-9}
Real answers shown & none & other pollsters & week before & same survey & none & other pollsters & week before & same survey \\
\midrule
Claude-Haiku-5.5 & 13 & 14 & 16 & 18 & 11 & 13 & 16 & 14 \\
GPT-5.6-luna & 15 & 16 & 15 & 19 & 14 & 13 & 13 & 13 \\
GPT-6.1-sol & 24 & 27 & 27 & 49 & 25 & 30 & 31 & 40 \\
GPT-6-astra & 26 & 23 & 27 & 52 & 26 & 22 & 28 & 43 \\
Claude-Opus-5.5 & 27 & 25 & 29 & 38 & 24 & 23 & 29 & 27 \\
Claude-Fable-5.2 & 30 & 31 & 33 & 54 & 29 & 34 & 34 & 40 \\
\bottomrule
\end{tabular}
\caption{Control experiment on the second through fifth weekly reports (206 questions; worth in respondents). Each prompt displays 29 real answers drawn from same-week U.S.\ polls by other pollsters, the same pollster's report from the prior week, or the other half of the same survey. Right: subset of 107 questions sharing no response options with the shown same-survey items.}
\label{tab:ctl}
\end{table}

\begin{table}[!htb]
\centering\small
\setlength{\tabcolsep}{5pt}
\begin{tabular}{lcccc}
\toprule
 & simulator & \multicolumn{2}{c}{real answers next to it} & real answers \\
 & alone & arithmetic & learned & in the prompt \\
\midrule
\multicolumn{5}{l}{\emph{A population's answer}} \\
GPT-5.6-luna & 15 & 18 & 19 & \textbf{19} \\
Claude-Haiku-5.5 & 13 & 14 & 15 & \textbf{18} \\
Claude-Opus-5.5 & 28 & 31 & 31 & \textbf{36} \\
GPT-6.1-sol & 24 & 32 & 27 & \textbf{47} \\
GPT-6-astra & 26 & 34 & 29 & \textbf{52} \\
Claude-Fable-5.2 & 31 & 38 & 34 & \textbf{54} \\
\midrule
\multicolumn{5}{l}{\emph{A group's answer} (overall result alone: 15)} \\
GPT-5.6-luna & 7 & 17 & 24 & \textbf{26} \\
Claude-Haiku-5.5 & 11 & 44 & 50 & \textbf{51} \\
Claude-Opus-5.5 & 18 & 81 & 90 & \textbf{128} \\
GPT-6.1-sol & 17 & 95 & 111 & \textbf{130} \\
GPT-6-astra & 18 & 101 & 121 & \textbf{130} \\
Claude-Fable-5.2 & 27 & 137 & 149 & \textbf{163} \\
\bottomrule
\end{tabular}
\caption{Comparison of ways to incorporate the same real answers on 246 questions (worth in respondents; group respondents in the lower panel). \emph{Arithmetic}: best of six post-hoc adjustments for populations and best of three for subgroups (chosen in-sample). \emph{Learned}: gradient-boosted trees trained out-of-sample on other questions. Population values reflect one random split.}
\label{tab:ways}
\end{table}

\paragraph{Post-hoc statistical corrections (Table~\ref{tab:ways}).} We compared in-prompt conditioning against three post-hoc baselines:
\begin{itemize}[leftmargin=*,itemsep=2pt,topsep=2pt]
\item \emph{Arithmetic corrections (population):} Six variants that recalibrate the cold answer on the shown questions, shift it by the simulator's mean error on shown items with the same response options (on either a linear or log scale, unweighted or weighted by text similarity), or combine both.
\item \emph{Arithmetic corrections (subgroup):} Three variants: (1) the additive anchor $\theta_{\text{all}}+(c_{\text{group}}-c_{\text{all}})$, (2) a linear combination of those three quantities with coefficients fitted on other questions, and (3) a multiplicative ratio adjustment $c_{\text{group}}\cdot(\theta_{\text{all}}/c_{\text{all}})$.
\item \emph{Learned correction:} Gradient-boosted trees predicting the log ratio of the real share to the cold prediction, taking as features the cold predictions, question and option text (24 TF-IDF components; dense sentence embeddings yield nearly identical results), and real shares of other items sharing the same options. Models are trained via 5-fold cross-validation grouped by question stem, with hyperparameters tuned on held-out stems. Prompting beats this learned correction for 12 of 13 simulators on full populations (by 1.6 to 1.8 times for GPT-6.1-terra, GPT-6.1-sol, GPT-6-astra, and Claude-Fable-5.2) and matches or beats it for all eleven GPT and Claude simulators on subgroups (by 1.0 to 3.0 times).
\end{itemize}
For the small-area comparison in Section~\ref{sec:real}, we combine each model estimate with $m=25$ respondents drawn from the group's published shares using the Fay--Herriot area-level weight $m/(m+r)$, where $r$ is estimated from the other cross-validation folds.

\paragraph{Prompting with other simulators' predictions.} To test whether other models' outputs can substitute for real survey answers in the prompt, we replaced the real half-survey results with the cold predictions of the two strongest models from the competing developer. Across the six tested simulators (three GPT, three Claude), this yields 26 to 32 respondents, never exceeding simple Equation~\ref{eq:set} pooling of those three cold predictions (26 to 33).

\paragraph{Thin surveys and strategic question selection.}
When every question already has a small human sample, placing all sample results (including the target question's) into the prompt yields 0.79 to 1.36 times the worth of simple Equation~\ref{eq:pool} pooling at 20 respondents per question across these six simulators, and 0.77 to 1.19 times at 50 respondents, with only GPT-5.6-luna falling below pooling at both sample sizes. Thus, in-prompt conditioning helps mainly when per-question samples are thin. With a budget of 50 interviews per question, asking every question and pooling is worth 63 to 81 respondents across the six simulators, against 30 to 72 for asking 100 people half of the questions (pooling on the asked half and prompting on the unasked half).

When a researcher must choose ten questions per report to field and simulate the rest, choosing diverse representative questions helps substantially. We test a simple greedy coverage rule that maximizes $\sum_i\max_{j\in S}s_{ij}$, where $s_{ij}$ is the TF-IDF cosine similarity between question texts $i$ and $j$. This uses no survey answers. With non-negative similarities and the empty-set value defined as zero, the coverage objective is monotone submodular, so greedy selection under the ten-question limit achieves a $1-1/e$ approximation guarantee \citep{nemhauser1978}. Treating asked questions as exact, greedy selection cuts total squared error by 10\% to 24\% relative to the mean of five random draws for five of the six simulators, beating every individual random draw for these five; for Claude-Haiku-5.5 it is 6\% lower than the mean but not below every draw.

\paragraph{Subgroup protocol and noise adjustment (Table~\ref{tab:grpci}).}
In the cold subgroup condition, each call lists all twenty groups and requests distributions for all respondents plus each group across four questions. In the shown condition, the prompt appends the published overall percentage next to each option. Because each published subgroup share $\hat\theta_g$ comes from a finite sample of $m$ group respondents (sampling variance $V/m$), we debias the observed MSE $\mathbb{E}\|\hat c_g-\hat\theta_g\|^2$ to recover the true MSE $\mathbb{E}\|\hat c_g-\theta_g\|^2$ by subtracting $V/m$ for cold estimates, and $(1-2\pi)V/m$ for estimates incorporating the observed overall result $\hat\theta_{\text{all}}$, where $\pi$ is the subgroup's share of the full sample. The latter holds exactly under unit sensitivity of $\hat c_g$ to $\hat\theta_{\text{all}}$ (since $\mathbb{E}\|\hat\theta_g-\theta_g\|^2-2\,\mathbb{E}\langle\hat c_g-\theta_g,\hat\theta_g-\theta_g\rangle=(1-2\pi)V/m$), and serves as a first-order approximation for in-prompt conditioning; omitting the adjustment leaves the main gains intact (3.6 to 9.3 times over cold, versus 3.7 to 10.0 with it).

If the published subgroup tables carry a survey design effect of 1.5 or 2, Claude-Fable-5.2's worth when shown the overall result rises from 163 to 178 or 196. Concretely, the weekly poll interviews roughly 170 Black and 200 Hispanic respondents; on these groups, Claude-Fable-5.2 given the overall result matches 103 and 118 group respondents. On exhaustive demographic partitions (sex, age, education, party), \emph{raking} (rescaling cold subgroup predictions so their population-weighted average matches the known overall share) reaches 33 to 169 respondents, versus 44 to 192 when the overall result is shown directly in the prompt. Table~\ref{tab:grpci} reports all eleven GPT and Claude simulators. Questions returned in the wrong format were re-queried one at a time with the number of entries fixed by the answer schema, and ten Claude-Sonnet-5.5 answers with shifted item IDs were aligned by position.

\begin{table}[!htb]
\centering\small
\setlength{\tabcolsep}{5pt}
\begin{tabular}{lcccccc}
\toprule
Simulator & asked cold & result + difference & free coefficients & result $\times$ ratio & shown the result & shown over cold \\
\midrule
GPT-6-luna & 6 {\scriptsize[5, 8]} & 11 {\scriptsize[9, 14]} & 16 {\scriptsize[13, 19]} & 10 {\scriptsize[8, 14]} & 36 {\scriptsize[29, 46]} & 6.0 {\scriptsize[4.7, 7.4]} \\
GPT-5.6-luna & 7 {\scriptsize[6, 9]} & 14 {\scriptsize[11, 17]} & 17 {\scriptsize[14, 20]} & 14 {\scriptsize[11, 18]} & 26 {\scriptsize[21, 32]} & 3.7 {\scriptsize[2.9, 4.5]} \\
Claude-Sonnet-5.5 & 9 {\scriptsize[7, 11]} & 16 {\scriptsize[13, 22]} & 20 {\scriptsize[17, 25]} & 16 {\scriptsize[13, 22]} & 87 {\scriptsize[76, 98]} & 10.0 {\scriptsize[7.6, 12.9]} \\
Claude-Haiku-5.5 & 11 {\scriptsize[8, 14]} & 40 {\scriptsize[32, 48]} & 42 {\scriptsize[34, 49]} & 44 {\scriptsize[35, 54]} & 51 {\scriptsize[42, 61]} & 4.7 {\scriptsize[3.6, 6.4]} \\
GPT-5.6-sol & 14 {\scriptsize[10, 20]} & 64 {\scriptsize[54, 75]} & 64 {\scriptsize[54, 74]} & 77 {\scriptsize[62, 94]} & 82 {\scriptsize[64, 99]} & 5.8 {\scriptsize[4.0, 8.1]} \\
GPT-6-sol & 15 {\scriptsize[11, 20]} & 66 {\scriptsize[56, 78]} & 67 {\scriptsize[57, 78]} & 75 {\scriptsize[62, 91]} & 88 {\scriptsize[75, 102]} & 5.9 {\scriptsize[4.3, 7.9]} \\
GPT-6.1-terra & 16 {\scriptsize[11, 22]} & 73 {\scriptsize[59, 89]} & 74 {\scriptsize[60, 90]} & 82 {\scriptsize[64, 103]} & 100 {\scriptsize[72, 130]} & 6.4 {\scriptsize[3.9, 9.4]} \\
GPT-6.1-sol & 17 {\scriptsize[12, 25]} & 81 {\scriptsize[71, 94]} & 87 {\scriptsize[77, 99]} & 95 {\scriptsize[80, 111]} & 130 {\scriptsize[116, 144]} & 7.7 {\scriptsize[5.2, 11.2]} \\
GPT-6-astra & 18 {\scriptsize[12, 27]} & 84 {\scriptsize[73, 95]} & 91 {\scriptsize[80, 102]} & 101 {\scriptsize[88, 116]} & 130 {\scriptsize[115, 147]} & 7.4 {\scriptsize[4.7, 11.1]} \\
Claude-Opus-5.5 & 18 {\scriptsize[13, 27]} & 74 {\scriptsize[56, 97]} & 78 {\scriptsize[60, 100]} & 81 {\scriptsize[58, 113]} & 128 {\scriptsize[112, 150]} & 7.1 {\scriptsize[4.8, 9.8]} \\
Claude-Fable-5.2 & 27 {\scriptsize[20, 37]} & 116 {\scriptsize[101, 132]} & 123 {\scriptsize[107, 139]} & 137 {\scriptsize[119, 157]} & 163 {\scriptsize[142, 184]} & 6.0 {\scriptsize[4.3, 8.4]} \\
\bottomrule
\end{tabular}
\caption{Subgroup worth across twenty groups and 246 questions (in group respondents, with 95\% intervals). \emph{Result + difference}: additive anchor (real overall result plus cold group-minus-everyone gap). \emph{Free coefficients}: linear combination of those three terms fitted out-of-sample. \emph{Result $\times$ ratio}: cold group prediction multiplied by the ratio of real to simulated overall shares. The overall result alone is worth 15 [13, 18].}
\label{tab:grpci}
\end{table}

\section{Groups in Two Further Datasets}
\label{app:groups}

To confirm that our subgroup findings are not specific to one pollster, we replicated them on two additional datasets: party-identification splits from the new polls (113 questions, 339 group answers) and twenty demographic/political groups from the older General Social Survey (36 questions, 720 group answers). Results appear in Table~\ref{tab:groups}.

\begin{table}[!htb]
\centering\small
\setlength{\tabcolsep}{3.2pt}
\begin{tabular}{lcccccccc}
\toprule
 & \multicolumn{6}{c}{Party groups, new polls} & \multicolumn{2}{c}{20 groups, GSS} \\
\cmidrule(lr){2-7}\cmidrule(lr){8-9}
Simulator & cold & pooled & +difference & separate calls & shown & $\rho$ & cold & +difference \\
\midrule
OLMo-3-7B & 3 {\scriptsize[2, 4]} & 11 {\scriptsize[9, 13]} & 6 {\scriptsize[5, 7]} & 6 & 3 {\scriptsize[3, 4]} & 0.68 & 6 {\scriptsize[4, 7]} & 34 {\scriptsize[28, 42]} \\
Mistral-Small-24B & 6 {\scriptsize[5, 6]} & 11 {\scriptsize[10, 13]} & 13 {\scriptsize[11, 17]} & 11 & 14 {\scriptsize[12, 16]} & 0.77 & 8 {\scriptsize[6, 10]} & 28 {\scriptsize[24, 34]} \\
Qwen3.5-9B & 4 {\scriptsize[3, 5]} & 11 {\scriptsize[9, 13]} & 7 {\scriptsize[6, 10]} & 5 & 6 {\scriptsize[5, 8]} & 0.69 & 5 {\scriptsize[4, 8]} & 40 {\scriptsize[34, 48]} \\
Gemma-4-12B & 5 {\scriptsize[4, 6]} & 11 {\scriptsize[10, 14]} & 9 {\scriptsize[8, 11]} & 7 & 9 {\scriptsize[7, 12]} & 0.68 & 8 {\scriptsize[6, 11]} & 28 {\scriptsize[23, 34]} \\
Qwen3.8-27B & 6 {\scriptsize[5, 8]} & 13 {\scriptsize[11, 15]} & 15 {\scriptsize[11, 21]} & 12 & 20 {\scriptsize[15, 28]} & 0.75 & 11 {\scriptsize[7, 18]} & 44 {\scriptsize[34, 57]} \\
GPT-6-luna & 8 {\scriptsize[6, 11]} & 13 {\scriptsize[10, 17]} & 17 {\scriptsize[10, 31]} & 14 & 36 {\scriptsize[28, 47]} & 0.74 & 23 {\scriptsize[18, 30]} & 103 {\scriptsize[87, 125]} \\
GPT-5.6-luna & 8 {\scriptsize[6, 10]} & 13 {\scriptsize[10, 16]} & 16 {\scriptsize[10, 25]} & 14 & 24 {\scriptsize[18, 33]} & 0.69 & 21 {\scriptsize[16, 29]} & 108 {\scriptsize[89, 133]} \\
Claude-Haiku-5.5 & 13 {\scriptsize[11, 15]} & 19 {\scriptsize[16, 23]} & 41 {\scriptsize[29, 58]} & 22 & 55 {\scriptsize[46, 66]} & 0.81 & 22 {\scriptsize[17, 31]} & 93 {\scriptsize[77, 115]} \\
Claude-Sonnet-5.5 & 13 {\scriptsize[10, 16]} & 18 {\scriptsize[14, 23]} & 30 {\scriptsize[17, 52]} & 16 & 26 {\scriptsize[15, 61]} & 0.75 & 30 {\scriptsize[21, 50]} & 151 {\scriptsize[109, 211]} \\
GPT-5.6-sol & 17 {\scriptsize[15, 20]} & 22 {\scriptsize[19, 26]} & 57 {\scriptsize[49, 66]} & 33 & 79 {\scriptsize[67, 94]} & 0.83 & 33 {\scriptsize[22, 57]} & 159 {\scriptsize[125, 204]} \\
GPT-6-sol & 20 {\scriptsize[15, 26]} & 24 {\scriptsize[19, 30]} & 62 {\scriptsize[48, 78]} & 37 & 83 {\scriptsize[66, 100]} & 0.82 & 45 {\scriptsize[31, 68]} & 170 {\scriptsize[131, 219]} \\
GPT-6.1-terra & 19 {\scriptsize[15, 25]} & 24 {\scriptsize[19, 29]} & 62 {\scriptsize[49, 80]} & 43 & 85 {\scriptsize[71, 107]} & 0.83 & 61 {\scriptsize[47, 81]} & 212 {\scriptsize[181, 255]} \\
Claude-Opus-5.5 & 25 {\scriptsize[20, 34]} & 29 {\scriptsize[25, 38]} & 73 {\scriptsize[58, 97]} & 53 & 96 {\scriptsize[78, 125]} & 0.81 & 110 {\scriptsize[80, 162]} & 352 {\scriptsize[300, 429]} \\
GPT-6.1-sol & 22 {\scriptsize[18, 28]} & 29 {\scriptsize[24, 37]} & 60 {\scriptsize[46, 81]} & 45 & 101 {\scriptsize[79, 140]} & 0.78 & 61 {\scriptsize[41, 106]} & 231 {\scriptsize[195, 278]} \\
GPT-6-astra & 24 {\scriptsize[18, 32]} & 30 {\scriptsize[24, 41]} & 66 {\scriptsize[50, 90]} & 49 & 102 {\scriptsize[76, 146]} & 0.79 & 60 {\scriptsize[39, 100]} & 146 {\scriptsize[81, 269]} \\
Claude-Fable-5.2 & 35 {\scriptsize[26, 47]} & 39 {\scriptsize[31, 50]} & 99 {\scriptsize[75, 134]} & 65 & 128 {\scriptsize[94, 187]} & 0.80 & 169 {\scriptsize[125, 260]} & 491 {\scriptsize[411, 597]} \\
\bottomrule
\end{tabular}
\caption{Subgroup worth (in group respondents, with 95\% intervals and observed sampling noise removed). \emph{Cold}: prompted for the group. \emph{Pooled}: best fixed linear mixture of the overall result and cold prediction. \emph{+difference}: anchor adding the simulator's group-minus-everyone gap to the overall result. \emph{Separate calls}: same anchor when group and overall shares are queried in separate calls. \emph{Shown}: overall result placed in the prompt. $\rho$: error correlation between the subgroup and all respondents. The overall result alone is worth 11 [9, 13] on party groups and 36 [30, 45] on GSS.}
\label{tab:groups}
\end{table}

\paragraph{Party groups in new polls.} Each call asks for five questions across four targets simultaneously: all respondents, Democrats, Republicans, and Independents. Here the overall result alone is worth 11 group respondents. GPT and Claude simulators score 8 to 35 cold, jump to 16 to 99 with the additive anchor, and reach 24 to 128 when shown the overall result in the prompt. By contrast, the three smallest open models fail to leverage the overall result (scoring 3 to 9 when shown it, below the 11-respondent baseline), while Mistral-Small-24B and Qwen3.8-27B reach 14 and 20. Querying the subgroup and full population in separate calls substantially weakens the anchor for GPT and Claude models (dropping it to 14 to 65), indicating that both should be elicited jointly so their errors stay aligned.

\paragraph{General Social Survey.} On GSS waves from 2016 to 2022 (36 opinion items across 20 groups spanning seven demographic categories, with 268 to 5,182 effective respondents per group), the overall result alone is worth 36 group respondents. Cold GPT and Claude predictions are worth 21 to 169, rising to 93 to 491 under the additive anchor. Because these surveys predate all models, absolute worth may partly reflect memorization.

\section{An Earlier Wave of the Same Question}
\label{app:wave}

\paragraph{Short-interval weekly polls.} Forty-four questions appeared in more than one weekly report; we treat the latest wave as the target and the immediately preceding wave (4 to 33 days earlier, median 12 days) as the real-data source. The earlier wave alone is worth 286 respondents [180, 486]. Cold simulators are worth far less, from 18 (Claude-Haiku-5.5) to 86 (Claude-Fable-5.2), and pooling any simulator with the earlier wave barely moves it (286 to 288). When six simulators are instead shown the earlier result together with its date and asked to predict the later wave, they adjust the prior shares by only 0.7 to 1.1 percentage points on average, yielding worths from 261 (GPT-5.6-luna) to 390 [287, 567] (Claude-Fable-5.2). Relative to the earlier answer alone, Claude-Fable-5.2 achieves a statistically significant gain of 1.36 times [1.04, 2.03], whereas the confidence intervals for all other tested simulators include one.

\paragraph{Multi-year GSS waves.} Over longer time horizons, earlier waves decay more, leaving slightly more room for pooling: on the GSS, using the 2016 and 2018 waves to predict 2021 and 2022 is worth 75 respondents [55, 118], and pooling them with the eleven GPT and Claude simulators raises worth to 85 to 171.

\end{document}